\documentclass[10pt]{article}
\usepackage{amsfonts}
\usepackage{fullpage,graphicx,subfigure,mathdots,mathpazo,color}
\usepackage{amsmath,amscd,tikz,mathrsfs,cite}
\usepackage[normalem]{ulem}
\usepackage{amsmath}
\usepackage{setspace}
\usepackage{epsfig,amsmath,graphicx,amssymb,overpic}
\usepackage{bm}
\usepackage{graphicx}
\usepackage{subfigure, xcolor}
\usepackage{ntheorem}
\usepackage{listings,diagbox}
\usepackage{color}
\usepackage{epsfig,amsmath,graphicx,amssymb,overpic}
\usepackage{booktabs}
\usepackage{diagbox}
\usepackage{graphicx}
\usepackage{dcolumn}
\usepackage{bm}
\usepackage{graphicx}
\usepackage{subfigure}
\usepackage{epsfig,amsmath,graphicx,amssymb,overpic,cite}
\usepackage{makecell}
\usepackage{fullpage,graphicx,subfigure,mathdots,mathpazo,color}
\usepackage{amsmath,amscd,tikz,mathrsfs,cite}
\usepackage[normalem]{ulem}
\usepackage{amsmath}
\usepackage{setspace}
\usepackage{epsfig,amsmath,graphicx,amssymb,overpic}
\usepackage{bm}
\usepackage{graphicx}
\usepackage{subfigure, xcolor}
\usepackage{ntheorem}

\def\be{\begin{equation}}
\def\ee{\end{equation}}
\def\bee{\begin{eqnarray}}
\def\ene{\end{eqnarray}}
\def\bes{\begin{subequations}}
\def\ees{\end{subequations}}

\newcommand{\br}{{\bf r}}

\newcommand{\PT}{{\cal PT}}

\newcommand{\bx}{{\bm x}}

\def\v{\vspace{0.1in}}

\allowdisplaybreaks[4]

\begin{document}

\baselineskip=13pt \renewcommand {\thefootnote}{\dag}
\renewcommand
{\thefootnote}{\ddag} \renewcommand {\thefootnote}{ }

\pagestyle{plain}




\baselineskip=14pt \renewcommand {\thefootnote}{\dag}
\renewcommand
{\thefootnote}{\ddag} \renewcommand {\thefootnote}{ }

\pagestyle{plain}

\begin{center}
\baselineskip=16pt \leftline{} \vspace{-.3in} {\Large \textbf{Data-driven 2D stationary quantum droplets and wave propagations in the amended GP equation with two potentials via deep neural networks learning}} \\[0.2in]

Jin Song$^{1,2}$ and Zhenya Yan$^{{1,2},{*}}$%
\footnote{$^{*}$Corresponding author. \textit{Email address}:
zyyan@mmrc.iss.ac.cn} \\[0.15in]
\textit{{\small $^{1}$KLMM, Academy of Mathematics and Systems Science,
Chinese Academy of Sciences, Beijing 100190, China \\[0pt]
$^{2}$School of Mathematical Sciences, University of Chinese Academy of
Sciences, Beijing 100049, China}} \\[0pt]
\end{center}

\v

\noindent \textbf{Abstract:}\thinspace\ In this paper, we develop a systematic deep learning approach to solve two-dimensional (2D) stationary quantum droplets (QDs) and investigate their wave propagation in the 2D amended Gross–Pitaevskii equation with Lee-Huang-Yang correction and two kinds of potentials. Firstly, we use the initial-value iterative neural network (IINN) algorithm for 2D stationary quantum droplets of stationary equations. Then the learned stationary QDs are used as the initial value conditions for physics-informed neural networks (PINNs) to explore their evolutions in the some space-time region. Especially, we consider two types of potentials, one is the 2D quadruple-well Gaussian potential and the other is the $\PT$-symmetric HO-Gaussian potential, which lead to spontaneous symmetry breaking and the generation of multi-component QDs. The used deep learning method can also be applied to study wave propagations of other nonlinear physical models.



\vspace{-0.05in} 

\baselineskip=13pt

\section{Introduction}\label{sec1}

Recent intensive researches have focused on quantum droplets (QDs), a new state of liquid matter~\cite{1}. QDs are characterized by a delicate balance between mutual attraction and repulsion, leading to their unique properties. QDs have potential applications in ultracold atoms and superfluids and have been studied widely \cite{3,4,5,6,7,8,9,10}. As the ultra-dilute liquid matter, QDs are nearly incompressible, self-sustained liquid droplets, with distinctive properties such as extremely low densities and temperatures \cite{11,12,13,14}. The Lee-Huang-Yang (LHY) effect \cite{15}, driven by quantum fluctuations, has been introduced to prevent QDs from collapsing due to mean-field approximation, enabling the prediction of stable QDs in weakly interacting Bose-Einstein condensates (BECs) \cite{1,4}.

Experimental realizations of QDs have been achieved in various systems, including single-component dipolar bosonic gases, binary Bose-Bose mixtures of different atomic states in $^{39}$K, and in the heteronuclear mixture of $^{41}$K and $^{87}$Rb atoms \cite{11,12,13,ob1,ob2}. The accurate description of QDs has been made possible by the amended Gross-Pitaevskii (GP) equation with Lee-Huang-Yang (LHY) correction, which has been shown to agree with experimental observations \cite{22,23}.
The reduction of dimensionality from 3D to 2D has a significant impact on the form of the LHY term. In this case, the repulsive quartic nonlinearity is replaced by a cubic nonlinearity with an additional logarithmic factor \cite{8}
such that the 2D amended GP equation in the binary BECs with two mutually symmetric components trapped in a potential can be written as the following dimensionless form after scaling
\begin{equation}\label{gp}
   i\psi_t=-\dfrac{1}{2}\nabla_{\br}^2\psi +2\ln(2|\psi|^2)|\psi|^2\psi+U(\br)\psi,
\end{equation}
where the complex wave function $\psi=\psi(\br, t)$,
${\br}=(x,y)$ stands for the 2D rescaled coordinates and $t\in\mathbb{R}$, $\nabla_{\br}^2=\partial^2/\partial x^2+\partial^2/\partial y^2$, $U(\br)$ is an external potential, which can be real or complex.
A variety of trapping configurations in BECs have allowed for the direct observation of fundamental manifestations of QDs. For instance, stable 2D anisotropic vortex QDs have been predicted in effectively 2D dipolar BECs \cite{Li}.  More importantly, vortical QDs have been found to be stable without the help of any potential by a systematic numerical investigation and analytical estimates \cite{liyy}. Additionally, vortex-carrying QDs can be experimentally generated in systems with attractive inter-species and repulsive intra-species interactions, confined in a shallow harmonic trap with an additional repulsive Gaussian potential at the center \cite{25}.
Furthermore, the exploration of QDs trapped in $\PT$-symmetric potentials has also been pursued \cite{26,27,28,29}.

Recently, there has been a surge in the development of deep neural networks for studying partial differential equations (PDEs). Various approaches, such as physics-informed neural networks (PINNs) \cite{pinn,pinn1,pinn2,pinn3,deepxde}, deep Ritz method \cite{deepritz}, and PDE-net \cite{pnet,pnet1}, have been proposed to effectively handle PDE problems. Among them, the PINNs method incorporates the physical constraints into the loss functions, allowing the models to learn and represent the underlying physics more accurately \cite{pinn,deepxde}.
Moreover, these deep learning methods have been extended to solve a wide range of PDEs in various fields \cite{twostage,pinndq,yan1,yan2,yan3,yan4,yan5,yan6,chen1,li1}.

For the general 2D stationary QDs in the form $\Psi(x,y,t)=\phi(x,y)e^{-i\mu t}$, solving for $\phi(x,y)$ is an important problem because $\phi$ serves as an initial-value condition of PINNs.
In general, the traditional numerical methods were developed thus far to compute solitary waves, including the Petviashvili method, accelerated imaginary-time evolution (AITEM)
method, squared-operator iteration (SOM) method, and Newton-conjugate-gradient (NCG) method \cite{pet,it,49,51,52}. More recently,
we proposed a new deep learning approach called the initial-value iterative neural network (IINN) for solitary wave computations of many types of nonlinear wave euqations~\cite{IINN}, which offers a mesh-free approach by taking advantage of automatic differentiation, and could overcomes the curse of dimensionality.

Motivated by the aforementioned discussions, the main objective of this paper is to develop a systematic deep learning approach to solve 2D stationary QDs, and investigate their evolutions in the amended Gross–Pitaevskii equation with potentials. Especially, we consider two types of potentials, one is 2D quadruple-well Gaussian potential and the other is $\PT$-symmetric HO-Gaussian potential, which lead to spontaneous symmetry breaking and the generation of multi-component QDs.
The remainder of this paper is arranged as follows. Firstly, we introduce the PINNs deep learning
framework for the evolution of QDs. And then the IINN framework for stationary QDs is presented in Sec 2.
In Sec. 3, data-driven 2D QDs in the amended GP equation with two types of potential are exhibited, respectively. Finally, we give some conclusions and discussions in Sec. 4.

\section{The framework of  deep learning method}

In the following, we focus on the trapped stationary QDs to Eq.~(\ref{gp}) in the form $\psi(\br, t) = \phi(\br)e^{-i\mu t}$, where $\mu$ stands for the chemical potential~\cite{1}, and $\lim_{|{\br}|\rightarrow \infty}\phi({\br})=0$ for $\phi({\br})\in\mathbb{R}[{\br}]$. Substituting the stationary solution into Eq.~(\ref{gp}) yields the following nonlinear stationary equation obeyed by the nonlinear localized eigenmode $\phi(\br)$:
\begin{equation}\label{2ds}
  \mu \phi=-\frac{1}{2}\nabla^2_{\br}\phi+2\ln(2|\phi|^2)|\phi|^2\phi+U(\br)\phi.
\end{equation}
In general, it is difficult to get the explicit, exact solutions of Eq.~(\ref{2ds}) with the potentials. For general parametric conditions, one can usually use the numerical iterative methods to solve Eq.~(\ref{2ds}) with zero-boundary conditions by choosing the proper initial value, such as Newton-conjugate-gradient (NCG) method~\cite{49}, the spectral renormalization method~\cite{50}, and the squared-operator iteration method~\cite{51}.
In this paper, we extend the deep learning IINN method for the computations of stationary QDs, and then we use the stationary QDs as the initial data to analyze the evolutions of QDs with the aid of PINNs.

\subsection{The IINN framework for stationary QDs}

Based on traditional numerical iterative methods and physics-informed neural networks (PINNs), recently we proposed the initial value iterative neural network (IINN) algorithm for solitary wave computations \cite{IINN}.
In the following, we will introduce the main idea of IINN method.
Two identical fully connected neural networks are employed to learn the desired solution $\phi^*$ of Eq.~(\ref{2ds}).

For the first network, we choose an appropriate initial value $\phi_0$ such that it is sufficiently close to $\phi^*$. Then we randomly select $N$ training points
$\{\br_i\}_{i=1}^{N}$ within the region and train the network parameters $\theta$ by minimizing the mean squared error loss $\mathcal{L}_1$, aiming to make the output of network $\bar{\phi}$ sufficiently close to initial value $\phi_0$, where loss function $\mathcal{L}_1$ is defined as follows
\begin{equation}\label{Loss1}
  \mathcal{L}_1:=\frac{1}{N}\sum_{i=1}^{N}|\bar{\phi}(\br_i)-\phi_0(\br_i)|^2.
\end{equation}

For the second network, we initialize the network parameters $\theta=\{W,B\}$ with the learned weights and biases from the first network,
that is
\begin{equation}\label{argmin}
  \theta_0=\mathrm{argmin}\, \mathcal{L}_1(\theta).
\end{equation}
For the network output $\hat{\phi}$,
we define the loss function $\mathcal{L}_2$ as follows and utilize SGD or Adam optimizer to minimize it.
\begin{equation}\label{Loss2}
  \mathcal{L}_2:=\frac{1}{N}\frac{\sum_{i=1}^{N}|L\hat{\phi}(\br_i)|^2}{\max_i(|\hat{\phi}(\br_i)|)}.
\end{equation}
It should be noted that $\mathcal{L}_2$ is different from the loss function $\mathcal{L}_0$ defined in PINNs. Here we are not taking boundaries into consideration, instead we incorporate $\max(|\hat{\phi}|)$ to ensure that $\hat{\phi}$ does not converge to trivial solution.

\subsection{The PINNs framework for the evolution of QDs}

Base on the obtained the stationary QDs in Sec. 3.1, we utilize the PINNs deep learning framework \cite{pinn} to address the data-driven solutions of Eq.~(\ref{gp}). The core concept of PINNs involves training a deep neural network to satisfy the physical laws and accurately represent the solutions for various nonlinear partial differential equations. In the case of the 2D amended GP equation (\ref{gp}), we incorporate initial-boundary value conditions.
\begin{equation}\label{ib}
 \left\{\begin{array}{l}
 i\psi_t+\dfrac{1}{2}\nabla_{\br}^2\psi -2\ln(2|\psi|^2)|\psi|^2\psi-U(\br)\psi=0,\quad
(\br, t)\in\Omega\times (0, T),\v\\ \psi(\br,0)=\phi(\br),\quad\br\in\Omega, \v\\
     \psi(\br,t)\big|_{\br\in \partial\Omega}=\phi_b(t),\quad t\in [0,T],
 \end{array}\right.
\end{equation}
where $\phi(\br)$ is the solution of stationary equation (\ref{2ds}), and we take $\phi_b(t)\equiv 0$, which is solved by the IINN method  in Sec. 3.1.

We rewrite the wave-function as $\psi(\br,t)=p(\br,t)+iq(\br,t)$ with $p(\br,t)$ and $q(\br,t)$ being its real and imaginary parts, respectively. Then the complex-valued PINNs $\mathcal{F}(\br, t)=\mathcal{F}_p(\br, t)+i\mathcal{F}_q(\br, t)$ with $\mathcal{F}_p(\br, t)$ and $\mathcal{F}_q(\br, t)$ being its real and imaginary parts, respectively can be defined as
\begin{equation}\label{F}
 \begin{array}{l}
     \displaystyle\mathcal{F}(\br, t):= i\psi_t+\dfrac{1}{2}\nabla_{\br}^2\psi -2\ln(2|\psi|^2)|\psi|^2\psi-U(\br)\psi,\v\\
     \displaystyle\mathcal{F}_p(\br, t):= -q_t+\dfrac{1}{2}\nabla^2_{\br}p-2\ln[2(p^2+q^2)](p^2+q^2)p-\mathrm{real}(U)p+\mathrm{imag}(U)q,\v\\
     \displaystyle\mathcal{F}_q(\br, t):= p_t+\dfrac{1}{2}\nabla^2_{\br}q-2\ln[2(p^2+q^2)](p^2+q^2)q-\mathrm{real}(U)q-\mathrm{imag}(U)p,
 \end{array}
\end{equation}
where $\mathrm{real}(U)$ and $\mathrm{imag}(U)$ represent the real and imaginary parts of the external potential $U(\br)$, respectively.
Therefore, a fully-connected neural network NN$(\br, t; W, B)$ with $i$ hidden Layers and $n$ neurons in every hidden layer can be constructed, where initialized parameters $W = \{w_j\}_{1}^{i+1}$ and $B = \{b_j\}_1^{i+1}$ being the weights and bias. Then, by the given activation function $\sigma$, one can obtain the expression in the form
\begin{equation}\label{sigma}
  A_j=\sigma(w_j\cdot A_{j-1}+b_j),
\end{equation}
where the $w_j$ is a dim$(A_j)\times $dim$(A_{j-1})$ matrix, $A_0$, $A_{i+1}$, $b_{i+1}\in \mathbb{R}^2$ and $A_j$, $b_j\in \mathbb{R}^n$.

Furthermore, a Python library for PINNs, DeepXDE, was designed to serve a research tool for solving problems in computational science and engineering \cite{deepxde}. Using DeepXDE, we can conveniently define the physics-informed neural network $\mathcal{F}(\br, z)$ as
\begin{lstlisting}
import deepxde as dde
def pde(x, psi):
    p = psi[:, 0:1]
    q = psi[:, 1:2]
    p_xx = dde.grad.hessian(psi, x, component=0, i=0, j=0)
    q_xx = dde.grad.hessian(psi, x, component=1, i=0, j=0)
    p_yy = dde.grad.hessian(psi, x, component=0, i=1, j=1)
    q_yy = dde.grad.hessian(psi, x, component=1, i=1, j=1)
    p_t = dde.grad.jacobian(psi, x, i=0, j=2)
    q_t = dde.grad.jacobian(psi, x, i=1, j=2)
    F_p = -q_t + 0.5*(p_xx+p_yy) - 2*tf.log(2*(p**2+q**2))*(p**2+q**2)*p - (V*p - W*q)
    F_q = p_t + 0.5*(q_xx+q_yy) - 2*tf.log(2*(p**2+q**2))*(p**2+q**2)*q - (V*q + W*p)
    return [F_p, F_q]
\end{lstlisting}

In order to train the neural network to fit the solutions of Eq.~(\ref{ib}), the total mean squared error (MSE) is defined as the following loss function containing three parts
\begin{equation}\label{loss}
  \mathcal{L}_0=MSE_F+MSE_I+MSE_B,
\end{equation}
with
\begin{equation}\label{lfib}
 \begin{array}{l}
     \displaystyle\quad MSE_F=\frac{1}{N_f}\sum_{\ell=1}^{N_f}\left(|\mathcal{F}_p(\br_f^\ell,t_f^\ell)|^2
     +|\mathcal{F}_q(\br_f^\ell,t_f^\ell)|^2\right),\v\\
     \displaystyle\quad MSE_I=\frac{1}{N_I}\sum_{\ell=1}^{N_I}\left(|p(\br_I^\ell,0)-p_0^\ell|^2
     +|q(\br_I^\ell,0)-q_0^\ell|^2\right),\v\\
     \displaystyle\quad MSE_B=\frac{1}{N_B}\sum_{\ell=1}^{N_B}\left(|p(\br_B^\ell,t_B^\ell)|^2
     +|q(\br_B^\ell,t_B^\ell)|^2\right),
 \end{array}
\end{equation}
where $\{\br_f^\ell,t_f^\ell\}_\ell^{N_f}$ are connected with the marked points in $\Omega\times[0,T]$ for the PINNs $\mathcal{F}(\br, t)=\mathcal{F}_p(\br, t)+i\mathcal{F}_q(\br, t)$,
$\{\br_I^\ell,p_0^\ell,q_0^\ell\}_\ell^{N_I}$ represent the initial data with $\phi(\br_I^\ell)=p_0^\ell+iq_0^\ell$, and $\{\br_B^\ell,t_B^\ell\}_\ell^{N_B}$ are linked with the randomly selected boundary training points in domain $\partial\Omega\times[0,T]$.

And then, we choose a hyperbolic tangent function $\tanh(\cdot)$ as the activation function (of course one can also choose other nonlinear functions as the activation functions), and use Glorot normal to initialize variate. Therefore, the fully connected neural network can be written in Python as follows
\begin{lstlisting}
data = dde.data.TimePDE(
    geomtime, pde,
    initial-boundary value conditions,
    N_f, N_B, N_I,
)
net = dde.maps.FNN(layer_size, "tanh", Glorot normal)
model = dde.Model(data, net)
\end{lstlisting}
And then, with the aid of some optimization approaches (e.g., Adam \& L-BFGS) \cite{adam,bfgs}, we minimize the whole MSE $\mathcal{L}_0$ to make the approximated solution satisfy Eq.~(\ref{gp}) and initial-boundary value conditions.
\begin{lstlisting}
model.compile("adam", lr=1.0e-3)
model.train(epochs)
model.compile("L-BFGS")
model.train()
\end{lstlisting}

Therefore, for the given initial condition $\phi(\br)$ solved by the IINNs, we can use PINNs to obtain solutions for the whole space-time region.

Therefore, the main steps by the combination of the IINN and PINNs deep learning methods to solve the amended GP equation (\ref{ib}) with potentials and the initial-boundary value conditions are presented as follows:
\begin{itemize}

\item [1)] Given an initial value that is sufficiently close to the stationary QDs we want to obtain. And a fully connected network NN$_1$ is trained to fit it.
Then the IINN method is used to solve Eq.~(\ref{2ds}).

\item [2)] We initialize the network parameters of the second network NN$_2$ with the learned weights and biases from
the first network NN$_1$, that is $\theta_0=\mathrm{argmin}\, \mathcal{L}_1(\theta)$. And train the NN$_2$ by minimizing the loss function $\mathcal{L}_2$ in terms of the optimization algorithm.

\item [3)] Construct a fully-connected neural network NN$(\br, t; \theta)$ with randomly initialized parameters, and the PINNs $\mathcal{F}(\br, t)$ is given by Eq.~(\ref{F}).

\item [4)] Generate the training data sets for the initial value condition given by IINN method, and considered model respectively from the initial boundary and within the region.

\item [5)] Construct a training loss function $\mathcal{L}_0$ given by Eq.~(\ref{loss}) by summing the MSE of both the $\mathcal{F}(\br, t)$ and initial-boundary value residuals.
And train the NN to optimize the parameters $\theta=\{W, B\}$ by minimizing the loss function in terms of
the Adam \& L-BFGS optimization algorithm.

\end{itemize}

In what follows, the deep learning scheme is used to investigate the data-driven QDs of the 2D amended GP equation (\ref{ib}) with two types of potential (quadruple-well Gaussian potential and $\PT$-symmetric HOG potential).

\section{Data-driven 2D QDs in amended GP equation with potentials}

\begin{figure*}[!t]
   \centering
\vspace{-0.15in}
  {\scalebox{0.80}[0.80]{\includegraphics{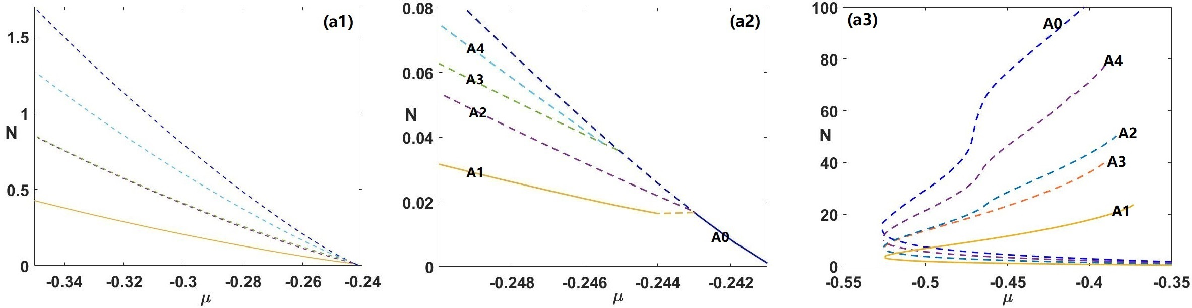}}}\hspace{-0.35in}
\vspace{0.15in}
\caption{\small (a1) Norm $N$ vs the chemical potential $\mu$ for 2D QDs in quadruple-well Gaussian potential (\ref{potential1}) starting from the ground state in the linear regime (dotted: unstable; solid: stable). (a2) Local diagrams of  relevant bifurcations in (a1). (a3)  Norm $N$ (big) vs the chemical potential $\mu$ for 2D QDs starting from the ground state in the linear regime including lower and upper branches~\cite{fourwell}.}
  \label{figssb1}
\end{figure*}

\subsection{Data-driven QDs in amended GP equation with quadruple-well Gaussian potential}

Firstly, we consider the 2D quadruple-well Gaussian potential~\cite{fourwell}
\begin{equation}\label{potential1}
  U(\br)= V_0\sum_{j=1}^4\exp\left[-k(\br-\br_j)^2 \right],\quad V_0<0,\quad k>0,
\end{equation}
where $\br_j=(\pm x_0,\pm y_0)$, $j=1,2,3,4$ control the locations of these four potential wells, and $|V_0|$ and $k$ regulate the depths and widths of potential wells, respectively. Recently, based on the usual numerical methods, the spontaneous symmetry breaking (SSB) of 2D QDs was considered for the amended GP equation with the potential (\ref{potential1})~\cite{fourwell}, in which the complete pitchfork symmetry breaking bifurcation diagrams were presented for the possible stationary states with four modes, which involve twelve different real solution branches and one complex solution branches (for the complex one, the norm $N=\int_{\mathbb{R}^2}|\phi|^2d^2{\bf r}$ is the same as for one real branch),
see Fig.~\ref{figssb1} for diagrams about the norm as a function of $\mu$, and stable/unstable modes.

In the following, we use the deep learning method to consider the four branches, that is, branches A0, A1, A3 and A4 (see Fig.~\ref{figssb1}).
It should be noted that for the same potential parameters and chemical potential $\mu$, Eq.~(\ref{2ds}) can admit different solutions, which cannot be solved by general deep learning methods.
Here we take potential parameters as $V_0 =-0.5$ and $k=0.1$, and consider $\Omega = [-12, 12]\times[-12,12]$, $T = 5$ and $\mu=-0.5$.
If not otherwise specified, we choose a 4-hidden-layer deep neural network with 100 neurons per layer, and set learning rate $\alpha = 0.001$.

\begin{figure*}[!t]
    \centering
  {\scalebox{0.80}[0.80]{\includegraphics{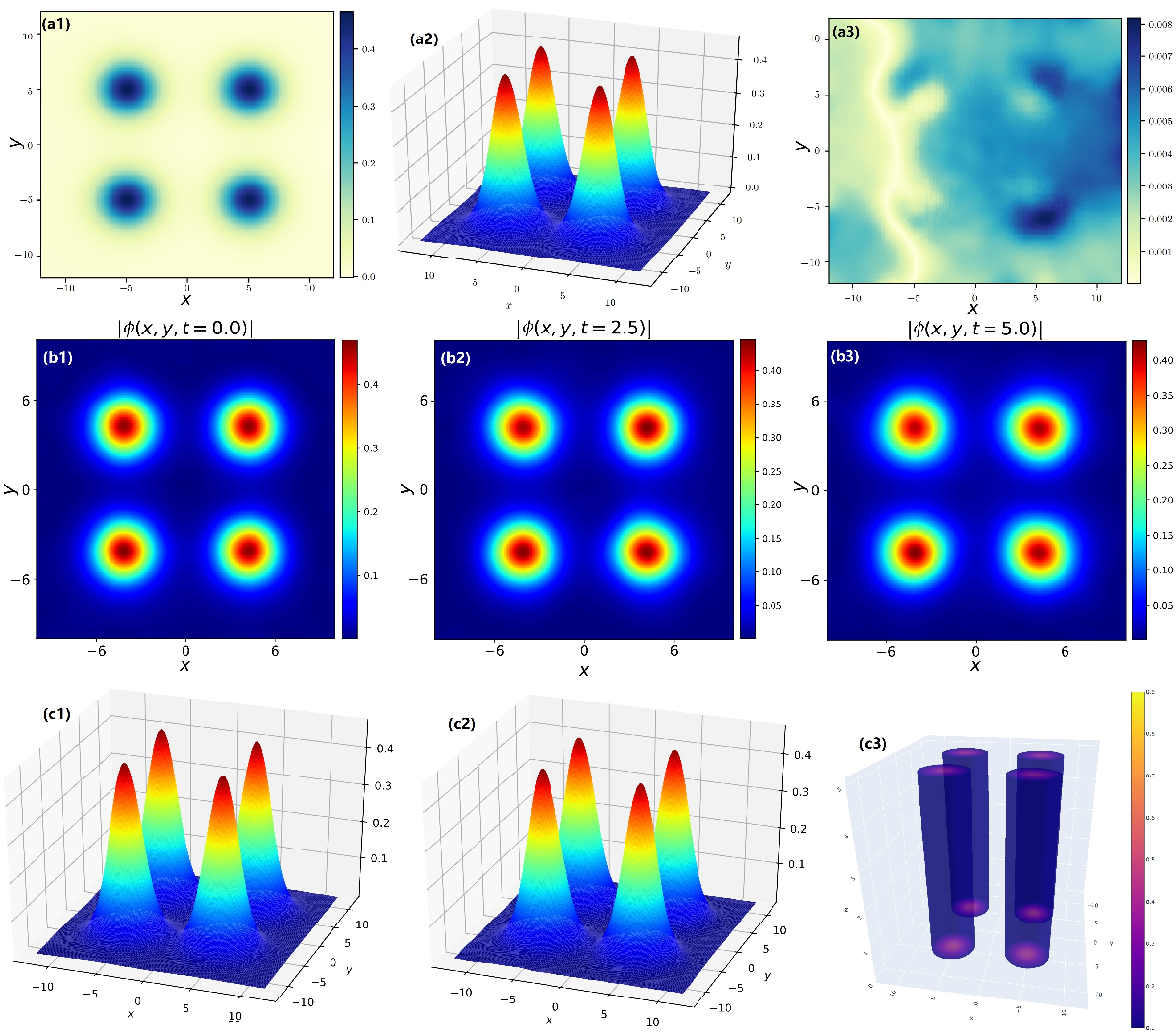}}}\hspace{-0.35in}
\vspace{0.15in}
\caption{\small The QDs in branch A0 of 2D amended GP equation with quadruple-well Gaussian potential. (a1) The intensity diagram $|\phi(\br)|$ of learned solution by IINN method.
(a2) The 3D profile of the learned solution. (a3) The module of absolute error between the exact and learned solutions.
(b1, b2, b3) The intensity diagram of the learned solutions at different time $t = 0,\, 2.5$, and $5.0$, respectively. (c1, c2) The initial state of the learned solution by IINN method and PINNs method. (c3) Isosurface of learned QDs at values 0.1, 0.5 and 0.9.}
  \label{fig1}
\end{figure*}

\begin{figure*}[!t]
    \centering
  {\scalebox{0.83}[0.83]{\includegraphics{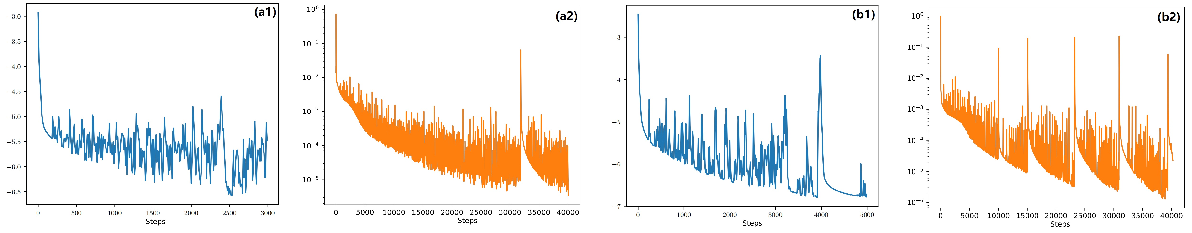}}}\hspace{-0.35in}
\vspace{0.15in}
\caption{\small The loss-iteration plots.
 (a1) The QDs in branch A0 for IINN. (a2) The QDs in branch A0 for PINNs.
  (b1) The QDs in branch A1 for IINN. (b2) The QDs in branch A1 for PINNs.
 }
  \label{lossfig}
\end{figure*}

\v{\it Case 1.}---In branch A0, we firstly obtain the stationary QDs by the IINN method. We set $N = 20000$, and take the initial value as
\begin{equation}\label{u01}
  \phi_0=\sum_{j=1}^4a_j\exp\left[-k(\br-\br_j)^2 \right],
\end{equation}
where $a_j=0.46\, (j=1,2,3,4)$ and $k=0.1$.
Through the IINN method, the learned QDs can be obtained at $\mu=-0.5$, whose intensity diagram $|\phi(\br)|$ and 3D profile
are shown in Figs.~\ref{fig1}(a1, a2), after 20000 steps of iterations with NN$_1$ and 3000 steps of iterations with NN$_2$. The relative $L_2$ error is 8.255472e-03 compared to the exact solution (numerically obtained). And the module of absolute error is exhibited in Fig.~\ref{fig1}(a3). The loss-iteration plot of NN$_1$ is displayed in Fig.~\ref{lossfig}(a1).

Then according to PINNs method, we take random sample points $N_f=20000$, $N_B=150$ and $N_I=1000$, respectively. Then, by using 40000 steps Adam and 10000 steps L-BFGS optimizations, we obtain the learned QDs solution $\hat{\psi}(\bx,t)$ in the whole space-time region.
Figs.~\ref{fig1}(b1, b2, b3) exhibit the magnitude of the predicted solution at different time $t = 0,\, 2.5$, and $5.0$, respectively. And the initial state ($\phi(\br)=\psi(\br,t=0)$) of the learned solution by IINN method and PINNs method is shown in Figs.~\ref{fig1}(c1, c2), respectively. Furthermore, nonlinear propagation simulation of the learned 2D QDs is displayed by the isosurface of learned soliton at values 0.1, 0.5 and 0.9 hereinafter (see Fig.~\ref{fig1}(c3)). The relative $L_2$ norm errors of $\psi(\br, t)$, $p(\br, t)$ and $q(\br, t)$, respectively, are 1.952e-02, 1.396e-02 and 1.061e-02. And the loss-iteration plot is displayed in Fig.~\ref{lossfig}(a2).

We should mention that the training stops in each step of the L-BFGS optimization when
\begin{equation}\label{stop}
  \frac{L_k-L_{k+1}}{\max\{|L_k|,|L_{k+1}|,1\}}\leq {\rm np.finfo(float).eps},
\end{equation}
where $L_k$ denotes loss in the n-th step L-BFGS optimization, and np.finfo(float).eps represent Machine Epsilon. Here we always set the default float type to `float64'. When the relative error between $L_k$ and $L_{k+1}$ is less than Machine Epsilon, the iteration stops.

\begin{figure*}[!t]
    \centering
  {\scalebox{0.80}[0.80]{\includegraphics{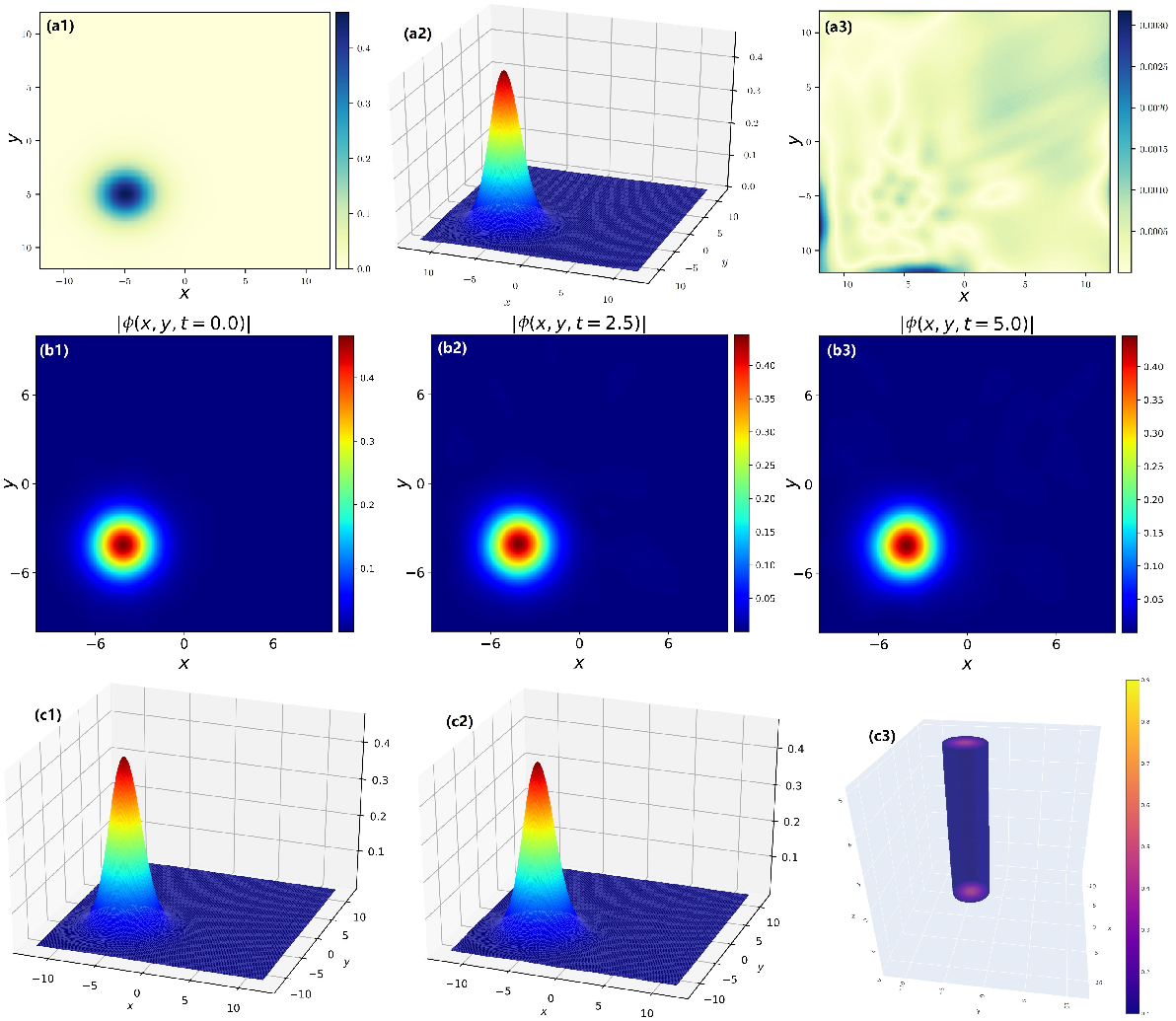}}}\hspace{-0.35in}
\vspace{0.15in}
\caption{\small The QDs in branch A1 of 2D amended GP equation with quadruple-well Gaussian potential. (a1) The intensity diagram $|\phi(\br)|$ of learned solution by IINN method.
(a2) The 3D profile of the learned solution. (a3) The module of absolute error between the exact and learned solutions.
(b1, b2, b3) The intensity diagram of the learned solutions at different time $t = 0,\, 2.5$, and $5.0$, respectively. (c1, c2) The initial state of the learned solution by IINN method and PINNs method. (c3) Isosurface of learned QDs at values 0.1, 0.5 and 0.9.}
  \label{fig2}
\end{figure*}

\v{\it Case 2.}---In branch A1, similarly we get the stationary QDs by IINN method. We set $N = 20000$, and take the initial value as
\begin{equation}\label{u02}
  \phi_0=\sum_{j=1}^4a_j\exp\left[-k(\br-\br_j)^2 \right],
\end{equation}
where $a_1=0.46,\, a_2=a_3=a_4=0$ and $k=0.1$.
According to the IINN method, the learned QDs can be obtained at $\mu=-0.5$, whose intensity diagram $|\phi(\br)|$ and 3D profile
are shown in Figs.~\ref{fig2}(a1, a2), after 10000 steps of iterations with NN$_1$ and 5000 steps of iterations with NN$_2$. The relative $L_2$ error is 8.821019e-03 compared to the exact solution. And the module of absolute error is exhibited in Fig.~\ref{fig2}(a3). The loss-iteration plot of NN$_1$ is displayed in Fig.~\ref{lossfig}(b1).

Then through the PINNs method, we take random sample points $N_f=20000$, $N_B=150$ and $N_I=1000$, respectively. Then, by using 40000 steps Adam and 10000 steps L-BFGS optimizations, we obtain the learned QDs solution $\hat{\psi}(\bx,t)$ in the whole space-time region.
Figs.~\ref{fig2}(b1, b2, b3) exhibit the magnitude of the predicted solution at different time $t = 0,\, 2.5$, and $5.0$, respectively. And the initial state ($\phi(\br)=\psi(\br,t=0)$) of the learned solution by IINN method and PINNs method is shown in Figs.~\ref{fig2}(c1, c2), respectively. Besides, nonlinear propagation simulation of the learned 2D QDs is displayed by the isosurface of learned soliton (see Fig.~\ref{fig2}(c3)). The relative $L_2$ norm errors of $\psi(\br, t)$, $p(\br, t)$ and $q(\br, t)$, respectively, are 3.364e-02, 3.767e-02 and 4.309e-02. And the loss-iteration plot is displayed in Fig.~\ref{lossfig}(b2).

\begin{figure*}[!t]
    \centering
  {\scalebox{0.80}[0.80]{\includegraphics{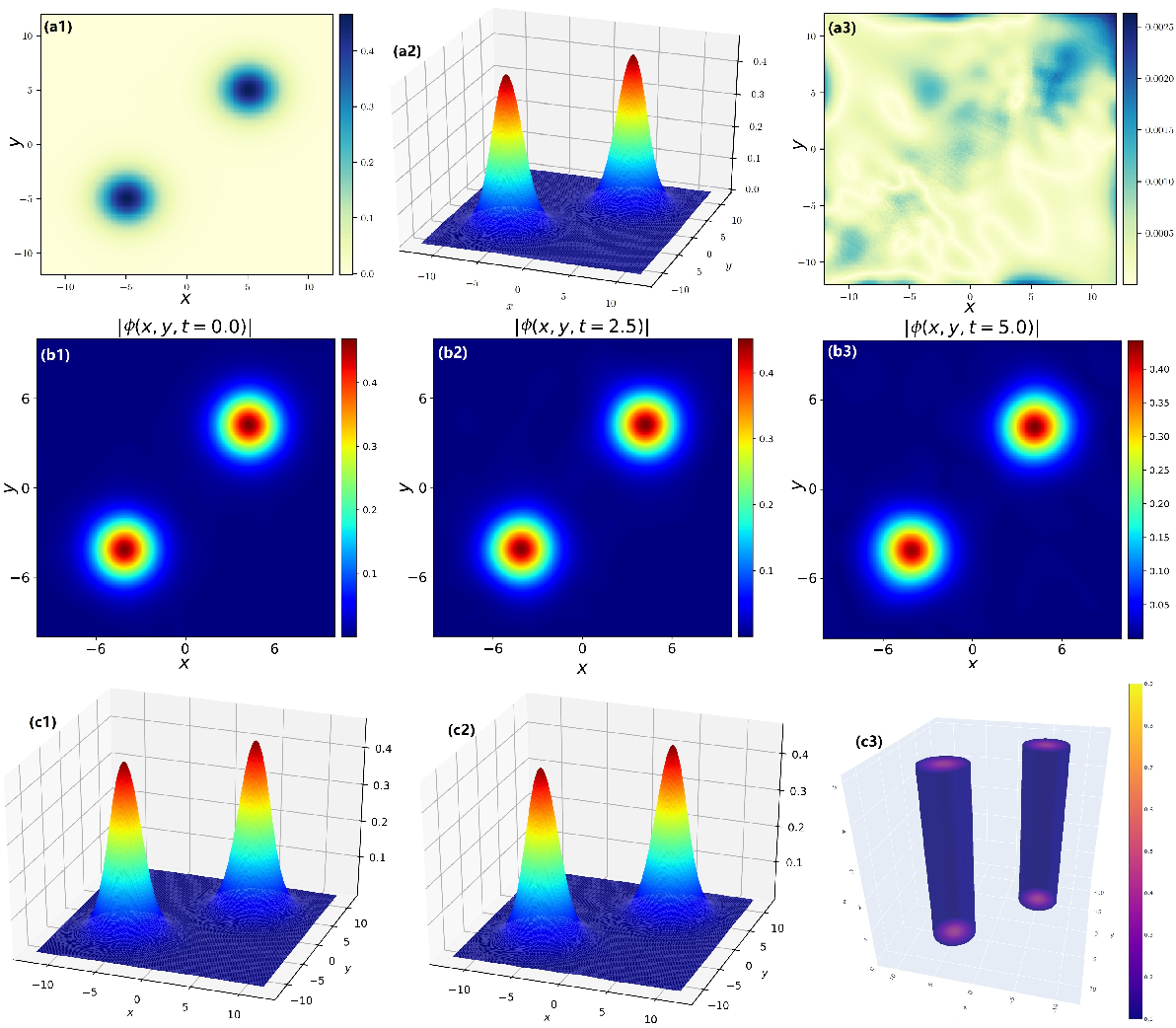}}}\hspace{-0.35in}
\vspace{0.15in}
\caption{\small The QDs in branch A3 of 2D amended GP equation with quadruple-well Gaussian potential. (a1) The intensity diagram $|\phi(\br)|$ of learned solution by IINN method.
(a2) The 3D profile of the learned solution. (a3) The module of absolute error between the exact and learned solutions.
(b1, b2, b3) The intensity diagram of the learned solutions at different time $t = 0,\, 2.5$, and $5.0$, respectively. (c1, c2) The initial state of the learned solution by IINN method and PINNs method. (c3) Isosurface of learned QDs at values 0.1, 0.5 and 0.9.}
  \label{fig3}
\end{figure*}

\v{\it Case 3.}---In branch A3, we firstly obtain the stationary QDs by IINN method. We set $N = 20000$, and take the initial value as
\begin{equation}\label{u03}
  \phi_0=\sum_{j=1}^4a_j\exp\left[-k(\br-\br_j)^2 \right],
\end{equation}
where $a_1=a_3=0.3,\, a_2=a_4=0$ and $k=0.1$.
Through the IINN method, the learned QDs can be obtained at $\mu=-0.5$, whose intensity diagram $|\phi(\br)|$ and 3D profile
are shown in Figs.~\ref{fig3}(a1, a2), after 15000 steps of iterations with NN$_1$ and 5000 steps of iterations with NN$_2$. The relative $L_2$ error is 7.201800e-03 compared to the exact solution. And the module of absolute error is exhibited in Fig.~\ref{fig3}(a3).

Then according to PINNs method, we take random sample points $N_f=20000$, $N_B=150$ and $N_I=1000$, respectively. Then, by using 40000 steps Adam and 10000 steps L-BFGS optimizations, we obtain the learned QDs solution $\hat{\psi}(\bx,t)$ in the whole space-time region.
Figs.~\ref{fig3}(b1, b2, b3) exhibit the magnitude of the predicted solution at different time $t = 0,\, 2.5$, and $5.0$, respectively. And the initial state ($\phi(\br)=\psi(\br,t=0)$) of the learned solution by IINN method and PINNs method is shown in Figs.~\ref{fig3}(c1, c2), respectively. Furthermore, nonlinear propagation simulation of the learned 2D QDs is displayed by the isosurface of learned soliton (see Fig.~\ref{fig3}(c3)). The relative $L_2$ norm errors of $\psi(\br, t)$, $p(\br, t)$ and $q(\br, t)$, respectively, are 2.002e-02, 2.458e-02 and 2.356e-02.

\begin{figure*}[!t]
    \centering
  {\scalebox{0.80}[0.80]{\includegraphics{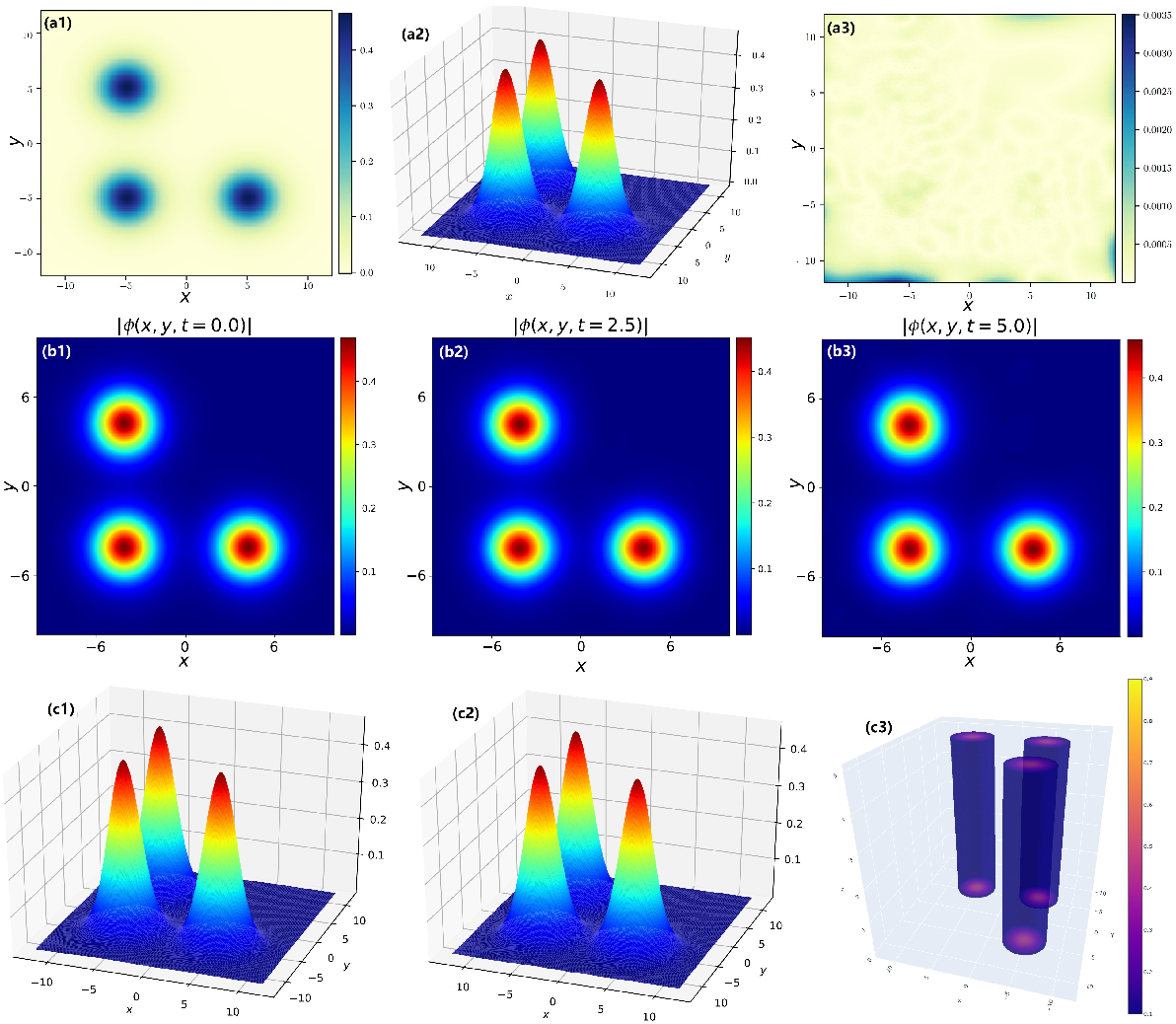}}}\hspace{-0.35in}
\vspace{0.15in}
\caption{\small The QDs in branch A4 of 2D amended GP equation with quadruple-well Gaussian potential. (a1) The intensity diagram $|\phi(\br)|$ of learned solution by IINN method.
(a2) The 3D profile of the learned solution. (a3) The module of absolute error between the exact and learned solutions.
(b1, b2, b3) The intensity diagram of the learned solutions at different time $t = 0,\, 2.5$, and $5.0$, respectively. (c1, c2) The initial state of the learned solution by IINN method and PINNs method. (c3) Isosurface of learned QDs at values 0.1, 0.5 and 0.9.}
  \label{fig4}
\end{figure*}

\v{\it Case 4.}---In branch A4, we get the stationary QDs by IINN method. We set $N = 20000$, and take the initial value as
\begin{equation}\label{u04}
  \phi_0=\sum_{j=1}^4a_j\exp\left[-k(\br-\br_j)^2 \right],
\end{equation}
where $a_1=a_2=a_3=0.46,\,a_4=0$ and $k=0.1$.
Through the IINN method, the learned QDs can be obtained at $\mu=-0.5$, whose intensity diagram $|\phi(\br)|$ and 3D profile
are shown in Figs.~\ref{fig4}(a1, a2), after 20000 steps of iterations with NN$_1$ and 3000 steps of iterations with NN$_2$. The relative $L_2$ error is 3.380430e-03 compared to the exact solution (numerically obtained). And the module of absolute error is exhibited in Fig.~\ref{fig4}(a3).

Then according to PINNs method, we take random sample points $N_f=20000$, $N_B=150$ and $N_I=1000$, respectively. Then, by using 40000 steps Adam and 10000 steps L-BFGS optimizations, we obtain the learned QDs solution $\hat{\psi}(\bx,t)$ in the whole space-time region.
Figs.~\ref{fig4}(b1, b2, b3) exhibit the magnitude of the predicted solution at different time $t = 0,\, 2.5$, and $5.0$, respectively. And the initial state ($\phi(\br)=\psi(\br,t=0)$) of the learned solution by IINN method and PINNs method is shown in Figs.~\ref{fig4}(c1, c2), respectively. Furthermore, nonlinear propagation simulation of the learned 2D QDs is displayed by the isosurface of learned soliton at values 0.1, 0.5 and 0.9 (see Fig.~\ref{fig4}(c3)). The relative $L_2$ norm errors of $\psi(\br, t)$, $p(\br, t)$ and $q(\br, t)$, respectively, are 1.256e-02, 1.899e-02 and 1.499e-02.

\begin{figure*}[!t]
   \centering
\vspace{-0.15in}
 {\scalebox{0.80}[0.80]{\includegraphics{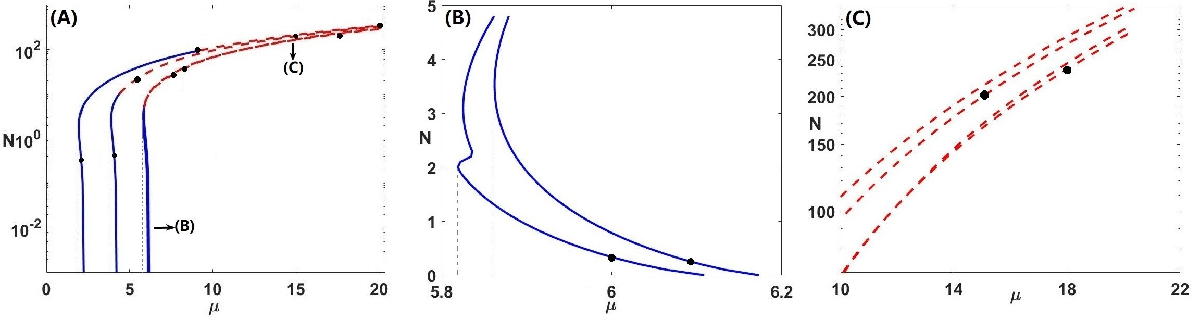}}}\hspace{-0.35in}
\vspace{0.15in}
\caption{\small (A) Norm $N$ vs the chemical potential $\mu$ for
different families of droplet modes in $\PT$-symmetric HOG potential (dashed: unstable; solid: stable). (B)
and (C) Zooms of the corresponding locations in (A)~\cite{spin}.} \label{figssb2}
\end{figure*}

\subsection{Data-driven QDs in amended GP equation with $\PT$-symmetric HOG potential}

In this subsection, we consider the following $\PT$-symmetric HO-Gaussian (HOG) potential with the real and imaginary parts being~\cite{spin}
\begin{equation}\label{potential2}
V(\br)=r^{2}\left( 1+e^{-r^{2}}\right) +V_{0}\,\left(e^{-2x^2}+e^{-2y^2}\right),\qquad
W(\br)=W_{0}\left( xe^{-x^{2}}+ye^{-y^{2}}\right) ,
\end{equation}%
where $r^{2}=x^{2}+y^{2}$, the coefficient in front of the HO potential is set to be 1,
the real parameter $V_{0}$ modulates the profile of the external potential $V(\br)$, and real $W_{0}$ is the strength of gain-loss distribution $W(\br)$. The vortex solitons were produced for a variety of 2D spinning QDs in the $\PT$-symmetric potential, modeled by the amended GP equation with Lee–Huang–Yang corrections~\cite{spin}, where the dependence of norm $N$ on chemical potential $\mu$ was illustrated for different families of droplet modes in $\PT$-symmetric HOG potential (see Fig.~\ref{figssb2}).

In the following, we use the deep learning method to consider the multi-component QDs under different chemical potential.
Here we take potential parameters as $V_0 =-1/16$ and $W_0=1$, and consider $\Omega = [-8, 8]\times[-8,8]$, $T = 3$.
Considering that the solution $\phi(\br)$ of Eq.~(\ref{2ds}) is a complex-valued function, similarly we set the network's output $\phi(\br)=p(\br)+iq(\br)$ and then separate Eq.~(\ref{2ds}) into its real and imaginary parts.
\begin{equation}\label{pq1}
 \begin{array}{l}
     \displaystyle\mathcal{F}_p(\br):= -\dfrac{1}{2}\nabla^2_{\br}p+2\ln[2(p^2+q^2)](p^2+q^2)p+\mathrm{real}(U)p-\mathrm{imag}(U)q-\mu p,\v\\
     \displaystyle\mathcal{F}_q(\br):= -\dfrac{1}{2}\nabla^2_{\br}q+2\ln[2(p^2+q^2)](p^2+q^2)q+\mathrm{real}(U)q+\mathrm{imag}(U)p-\mu q.
 \end{array}
\end{equation}
Then the loss function $\mathcal{L}_2$ becomes
\begin{equation}\label{L2new}
  \mathcal{L}_2:=\frac{1}{N}\frac{\sum_{i=1}^{N}\left(|\mathcal{F}_p(\br_i)|^2+|\mathcal{F}_q(\br_i)|^2\right)}{\max_i\left(\sqrt{(p(\br_i)^2+q(\br_i)^2}\right)}.
\end{equation}

\begin{figure*}[!t]
    \centering
  {\scalebox{0.80}[0.80]{\includegraphics{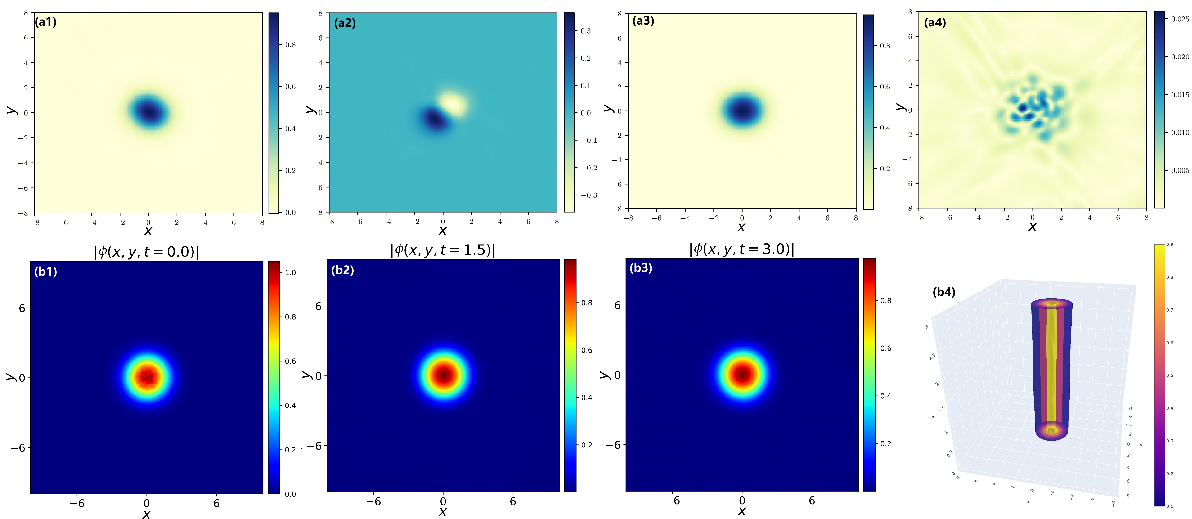}}}\hspace{-0.35in}
\vspace{0.15in}
\caption{\small Droplets with the one-component structure of 2D amended GP equation with $\PT$-symmetric HOG potential. (a1, a2, a3) The real part, imaginary part and intensity diagrams of
learned solution at $\mu= 2$. (a4) The module of absolute error between the exact and learned solutions.
(b1, b2, b3) The intensity diagram of the learned solutions at different time $t = 0,\, 1.5$, and $3.0$, respectively. (b4) Isosurface of learned QDs at values 0.1, 0.5 and 0.9.}
  \label{fig5}
\end{figure*}

\v{\it Case 1.\, QDs with the one-component structure.}---Firstly, we consider the $\PT$-symmetric droplets with the simplest structure.
We can obtain the initial conditions by computing the spectra and eigenmodes in the linear regime, which can be given as follows
\begin{equation}
\mathcal{H}\Phi (\mathbf{r})=\lambda \Phi (\mathbf{r}),\quad \mathcal{H}%
=-\nabla _{\mathbf{r}}^{2}+U(\br),\quad
\label{ls}
\end{equation}%
where $\lambda $ and $\Phi (\mathbf{r})$ are the eigenvalue and localized
eigenfunction, respectively. The linear spectral problem (\ref{ls}) can be solved
numerically by dint of the Fourier spectral method \cite{52}.

We take the initial value as the linear mode $\Phi$ at ground state and $N=10000$.
Through the IINN method, the learned QDs can be obtained at $\mu=2$, after 10000 steps of iterations with NN$_1$ and 10000 steps of iterations with NN$_2$.
Figs.~\ref{fig5}(a1, a2, a3) exhibit the intensity diagram of real part, imaginary part and $|\phi(\br)|$.  The module of absolute error is shown in Fig.~\ref{fig5}(a4).
The relative $L_2$ errors of $\phi(\br)$, $p(\br)$ and $q(\br)$, respectively, are 1.992564e-02, 5.547692e-02 and 2.075972e-02.

Then according to PINNs method, we take random sample points $N_f=20000$, $N_B=150$ and $N_I=1000$, respectively. Then, by using 30000 steps Adam and 10000 steps L-BFGS optimizations, we obtain the learned $\PT$ QDs solution $\hat{\psi}(\bx,t)$ in the whole space-time region.
Figs.~\ref{fig5}(b1, b2, b3) exhibit the magnitude of the predicted solution at different time $t = 0,\, 1.5$, and $3.0$, respectively. And, nonlinear propagation simulation of the learned 2D QDs is displayed by the isosurface of learned soliton at values 0.1, 0.5 and 0.9 hereinafter (see Fig.~\ref{fig5}(b4)). The relative $L_2$ norm errors of $\psi(\br, t)$, $p(\br, t)$ and $q(\br, t)$, respectively, are 2.312e-02, 1.995e-02 and 2.331e-02.

\begin{figure*}[!t]
    \centering
  {\scalebox{0.80}[0.80]{\includegraphics{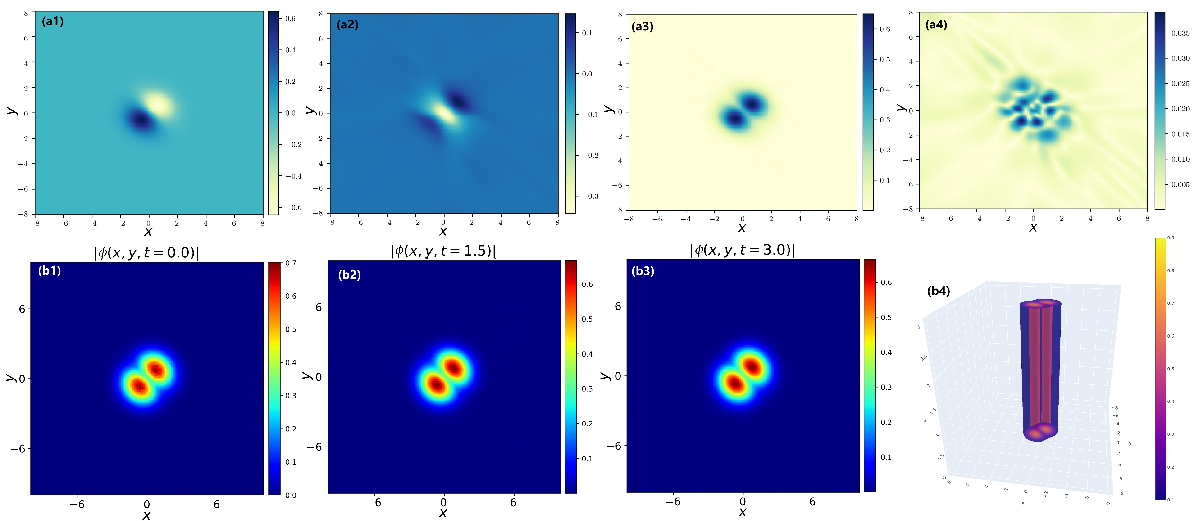}}}\hspace{-0.35in}
\vspace{0.15in}
\caption{\small Droplets with the two-component structure of 2D amended GP equation with $\PT$-symmetric HOG potential. (a1, a2, a3) The real part, imaginary part and intensity diagrams of
learned solution at $\mu= 2.8$. (a4) The module of absolute error between the exact and learned solutions.
(b1, b2, b3) The intensity diagram of the learned solutions at different time $t = 0,\, 1.5$, and $3.0$, respectively. (b4) Isosurface of learned QDs at values 0.1, 0.5 and 0.9.}
  \label{fig6}
\end{figure*}

\v{\it Case 2.\, QDs with the two-component structure}---Second, we consider the $\PT$-symmetric droplets with the two-component structure.
We take the initial value as the linear mode $\Phi$ at the first excited state and $N=10000$.
Through the IINN method, the learned QDs can be obtained at $\mu=2.8$, after 10000 steps of iterations with NN$_1$ and 10000 steps of iterations with NN$_2$.
Figs.~\ref{fig6}(a1, a2, a3) exhibit the intensity diagram of real part, imaginary part and $|\phi(\br)|$.  The module of absolute error is shown in Fig.~\ref{fig6}(a4).
The relative $L_2$ errors of $\phi(\br)$, $p(\br)$ and $q(\br)$, respectively, are 5.231775e-02, 1.546516e-02 and 6.117475e-02.

Then according to PINNs method, we take random sample points $N_f=20000$, $N_B=150$ and $N_I=1000$, respectively. Then, by using 30000 steps Adam and 15000 steps L-BFGS optimizations, we obtain the learned $\PT$ QDs solution $\hat{\psi}(\bx,t)$ in the whole space-time region.
Figs.~\ref{fig6}(b1, b2, b3) exhibit the magnitude of the predicted solution at different time $t = 0,\, 1.5$, and $3.0$, respectively. And, nonlinear propagation simulation of the learned 2D QDs is displayed by the isosurface of learned soliton at values 0.1, 0.5 and 0.9 hereinafter (see Fig.~\ref{fig6}(b4)). The relative $L_2$ norm errors of $\psi(\br, t)$, $p(\br, t)$ and $q(\br, t)$, respectively, are 4.326e-02, 5.158e-02 and 5.182e-02.

\begin{figure*}[!t]
    \centering
  {\scalebox{0.80}[0.80]{\includegraphics{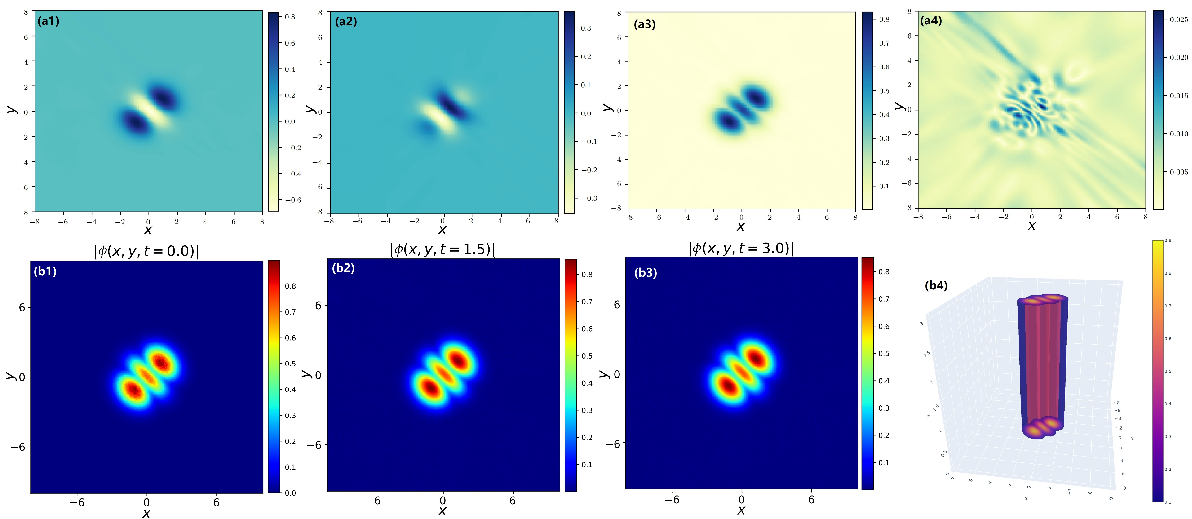}}}\hspace{-0.35in}
\vspace{0.15in}
\caption{\small Droplets with the three-component structure of 2D amended GP equation with $\PT$-symmetric HOG potential. (a1, a2, a3) The real part, imaginary part and intensity diagrams of
learned solution at $\mu= 4.3$. (a4) The module of absolute error between the exact and learned solutions.
(b1, b2, b3) The intensity diagram of the learned solutions at different time $t = 0,\, 1.5$, and $3.0$, respectively. (b4) Isosurface of learned QDs at values 0.1, 0.5 and 0.9.}
  \label{fig7}
\end{figure*}

\v{\it Case 3.\, QDs with the three-component structure}---Then, we consider the $\PT$-symmetric droplets with the three-component structure.
We take the initial value as the linear mode $\Phi$ at the second excited state and $N=10000$.
Through the IINN method, the learned QDs can be obtained at $\mu=4.3$, after 10000 steps of iterations with NN$_1$ and 10000 steps of iterations with NN$_2$.
Figs.~\ref{fig7}(a1, a2, a3) exhibit the intensity diagram of real part, imaginary part and $|\phi(\br)|$.  The module of absolute error is shown in Fig.~\ref{fig7}(a4).
The relative $L_2$ errors of $\phi(\br)$, $p(\br)$ and $q(\br)$, respectively, are 3.606203e-02, 5.148407e-02 and 4.234037e-02.

Then according to PINNs method, we take random sample points $N_f=20000$, $N_B=150$ and $N_I=1000$, respectively. Then, by using 30000 steps Adam and 15000 steps L-BFGS optimizations, we obtain the learned $\PT$ QDs solution $\hat{\psi}(\bx,t)$ in the whole space-time region.
Figs.~\ref{fig7}(b1, b2, b3) exhibit the magnitude of the predicted solution at different time $t = 0,\, 1.5$, and $3.0$, respectively. And, nonlinear propagation simulation of the learned 2D QDs is displayed by the isosurface of learned soliton at values 0.1, 0.5 and 0.9 hereinafter (see Fig.~\ref{fig7}(b4)). The relative $L_2$ norm errors of $\psi(\br, t)$, $p(\br, t)$ and $q(\br, t)$, respectively, are 5.410e-02, 6.771e-02 and 6.189e-02.

\begin{figure*}[!t]
    \centering
  {\scalebox{0.80}[0.80]{\includegraphics{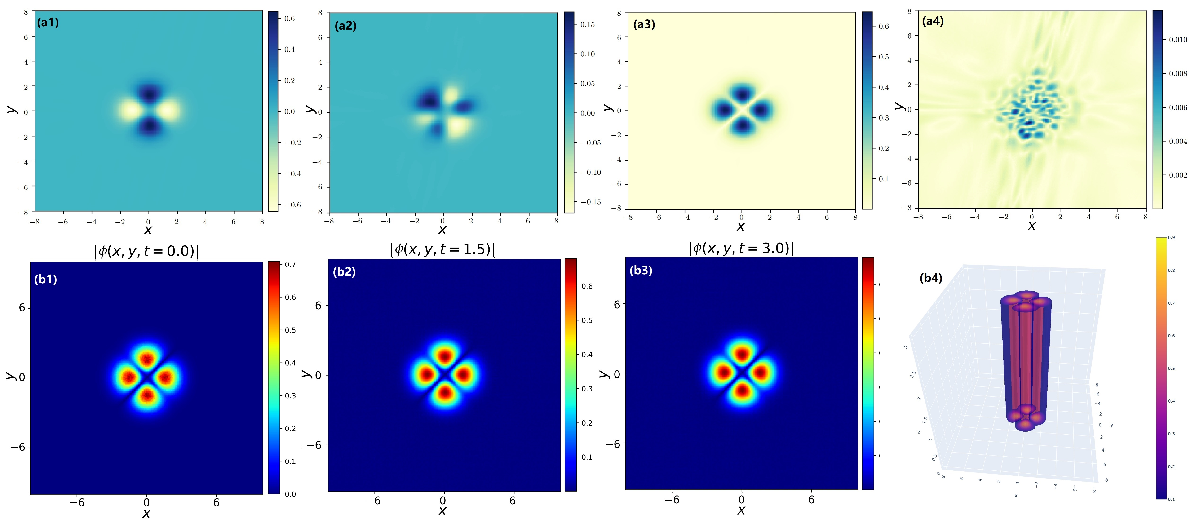}}}\hspace{-0.35in}
\vspace{0.15in}
\caption{\small Droplets with the four-component structure of 2D amended GP equation with $\PT$-symmetric HOG potential. (a1, a2, a3) The real part, imaginary part and intensity diagrams of
learned solution at $\mu= 4.2$. (a4) The module of absolute error between the exact and learned solutions.
(b1, b2, b3) The intensity diagram of the learned solutions at different time $t = 0,\, 1.5$, and $3.0$, respectively. (b4) Isosurface of learned QDs at values 0.1, 0.5 and 0.9.}
  \label{fig8}
\end{figure*}

\v{\it Case 4.\, QDs  with the four-component structure}---Finally, we consider the $\PT$-symmetric droplets with the four-component structure.
We take the initial value as the linear mode $\Phi$ at the three excited state and $N=10000$.
Through the IINN method, the learned QDs can be obtained at $\mu=4.2$, after 10000 steps of iterations with NN$_1$ and 20000 steps of iterations with NN$_2$.
Figs.~\ref{fig8}(a1, a2, a3) exhibit the intensity diagram of real part, imaginary part and $|\phi(\br)|$.  The module of absolute error is shown in Fig.~\ref{fig8}(a4).
The relative $L_2$ errors of $\phi(\br)$, $p(\br)$ and $q(\br)$, respectively, are 1.186940e-02, 4.332541e-02 and 1.185871e-02.

Then according to PINNs method, we take random sample points $N_f=20000$, $N_B=150$ and $N_I=1000$, respectively. Then, by using 30000 steps Adam and 15000 steps L-BFGS optimizations, we obtain the learned $\PT$ QDs solution $\hat{\psi}(\bx,t)$ in the whole space-time region.
Figs.~\ref{fig8}(b1, b2, b3) exhibit the magnitude of the predicted solution at different time $t = 0,\, 1.5$, and $3.0$, respectively. And, nonlinear propagation simulation of the learned 2D QDs is displayed by the isosurface of learned soliton at values 0.1, 0.5 and 0.9 hereinafter (see Fig.~\ref{fig8}(b4)). The relative $L_2$ norm errors of $\psi(\br, t)$, $p(\br, t)$ and $q(\br, t)$, respectively, are 5.128e-02, 5.841e-02 and 5.669e-02.

\begin{figure}[!t]
    \centering
  {\scalebox{0.55}[0.55]{\includegraphics{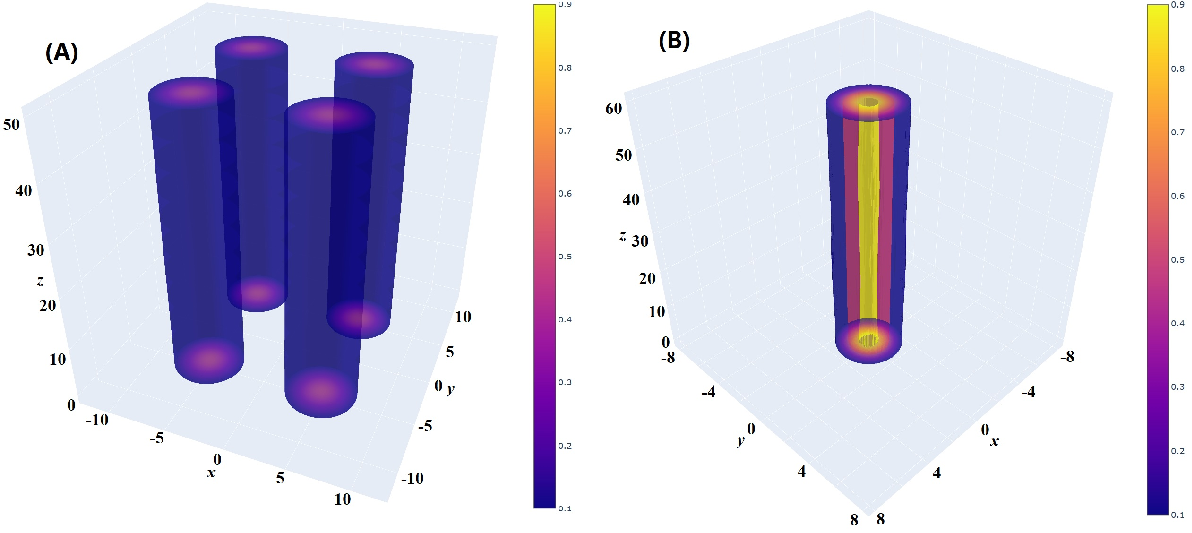}}}\hspace{-0.35in}
\vspace{0.15in}
\caption{\small  Longer-time evolutions for isosurface of learned QDs at values 0.1, 0.5 and 0.9. (A) The QDs in branch A0 of 2D amended GP equation with quadruple-well Gaussian potential (see Fig.~\ref{fig1}(c3)). (B) Droplets with the one-component structure of 2D amended GP equation with $\PT$-symmetric HOG potential (see Fig.~\ref{fig5}(b4)).}
  \label{figlong}
\end{figure}

 \v {\bf Remark.} It should be noted that the systematic results shown in Figs.~\ref{figssb1}and \ref{figssb2} have been investigated by numerical methods in Refs.~\cite{fourwell,spin}. In this paper, we mainly consider partial solutions and their short evolutions by using the machine learning method. For their stability, in general it can be solved by linear eigenvalue problems and long time evolution.   However, solving the eigenvalue problem via machine learning methods may be more difficult because it also involves multi-solution problems.  This will be our future work to consider.
  Furthermore, many methods already exist to return stable and accurate predictions across long temporal horizons by training multiple individual networks in different temporal sub-domains \cite{xPPINN,PPINN,ednn,longdeeponet}. These approaches inevitably lead to a large computational cost and more complex network structure. We use the parallel PINNs to investigate the longer-time evolutions for both solutions via the domain decomposition (see Fig.~\ref{figlong}). We can see that the solutions are still stable after the longer-time evolutions,  compared with Fig.~\ref{fig1}(c3) and Fig.~\ref{fig5}(b4), respectively. 

\section{Conclusions and discussions}

In conclusion, we have investigated the 2D stationary QDs and their evolutions in amended Gross–Pitaevskii equation with potentials via deep learning neural networks. Firstly, we use the IINN method for learning 2D stationary QDs. Then the learned 2D stationary QDs are used as the initial-value conditions for PINNs to display  their evolutions in the some space-time regions. Especially, we consider two types of potentials, one is 2D quadruple-well Gaussian potential and the other is $\PT$-symmetric HO-Gaussian potential, which lead to spontaneous symmetry breaking and the generation of multi-component QDs.

On the other hand, in order to study the stability of the QDs, we can use deep learning methods to study the interactions between droplets. Furthermore, we can investigate the spinning QDs in terms of spinning coordinates, $x^{\prime }=x\cos (\omega t)+y\sin (\omega t)$, $y^{\prime }=y\cos (\omega t)-x\sin (\omega t)$ with angular velocity $\omega$. We will investigate these issues in future.

 \v\v \noindent {\bf \large Acknowledgement}

\v The work  was supported by the National Natural Science Foundation of China  under Grant No. 11925108.


\end{document}